\documentclass[letterpaper, 10 pt, conference]{ieeeconf}  % Comment this line out if you need a4paper
\IEEEoverridecommandlockouts
 
\usepackage{cite}
\usepackage{amsmath,amssymb,amsfonts}
\newtheorem{definition}{Definition}
\newtheorem{theorem}{Theorem}
\newtheorem{assumption}{Assumption}
\newtheorem{lemma}{Lemma}

\usepackage{graphicx}
\usepackage{textcomp}
\usepackage{color}
\usepackage{subfigure}
\usepackage{multirow}
\usepackage{multicol}
\usepackage{makecell}
\usepackage{bm}
\usepackage{algorithm}
\usepackage{algpseudocode}
\usepackage{booktabs}
\usepackage{threeparttable}
\newcommand{\ie}{i.e.\ }

\newcommand{\Reffig}[1]{Figure~\ref{#1}}
\newcommand{\Refsec}[1]{Section~\ref{#1}}

\newcommand{\Refeq}[1]{Equation~\eqref{#1}}
\newcommand{\Reftab}[1]{Table~\ref{#1}}

\begin{document}
\title{
    Safe Reinforcement Learning with Dead-Ends Avoidance and Recovery
}

\author{Xiao Zhang,
    Hai Zhang,
    Hongtu Zhou,
    Chang Huang,
    Di Zhang,
    Chen Ye\textsuperscript{*},
    Junqiao~Zhao\textsuperscript{*}, ~\IEEEmembership{Member,~IEEE,}
    \thanks{*This work is supported by the National Key Research and Development Program of China (No. 2021YFB2501104, No. 2020YFA0711402)}
    \thanks{All the authors are with the Department of Computer Science and Technology, Tongji University, China and the MOE Key Lab of
        Embedded System and Service Computing, Tongji University, China, e-mail: (zhaojunqiao@tongji.edu.cn)}
}

\maketitle

\begin{abstract}
    Safety is one of the main challenges in applying reinforcement learning to realistic environmental tasks.
    To ensure safety during and after training process, existing methods tend to adopt overly conservative policy to avoid unsafe situations.
    However, overly conservative policy severely hinders the exploration, and makes the algorithms substantially less rewarding.
    In this paper, we propose a method to construct a boundary that discriminates safe and unsafe states.
    The boundary we construct is equivalent to distinguishing dead-end states, indicating the maximum extent to which safe exploration is guaranteed, and thus has minimum limitation on exploration.
    Similar to Recovery Reinforcement Learning, we utilize a decoupled RL framework to learn two policies, (1) a task policy that only considers improving the task performance, and (2) a recovery policy that maximizes safety.
    The recovery policy and a corresponding safety critic are pretrained on an offline dataset, in which the safety critic evaluates upper bound of safety in each state as awareness of environmental safety for the agent.
    During online training, a behavior correction mechanism is adopted, ensuring the agent to interact with the environment using safe actions only.
    Finally, experiments of continuous control tasks demonstrate that our approach has better task performance with less safety violations than state-of-the-art algorithms.
\end{abstract}

% \titlepgskip=-15pt

\section[sec:introduction]{INTRODUCTION}
\label{sec:introduction}

% \COMMENT{conservatism of the safety critic or the safe policy or the task policy, should be very careful used in descriptions}

% \COMMENT{safe / safety critic and policy should be consistent in descriptions}

Reinforcement learning (RL) has made impressive achievements in long-term control tasks, including Atria games\cite{mnih2015human}, car driving\cite{kendall2019learning} and robot controling\cite{bodnar2019quantile}.
While RL performs well in games and simulation environment, safety becomes one of the greatest challenge when applying RL to real-world task.
In real environments such as autonomous driving tasks, unsafe actions can lead to damage to the agent itself and to the environment, resulting in significant maintenance costs and even human casualties.
% As in the case of autonomous driving tasks, unsafe actions greatly increase the risk of car accidents, resulting in economic losses and even human casualties.

\begin{figure}[!ht]
    \centering
    \includegraphics[width=1\linewidth]{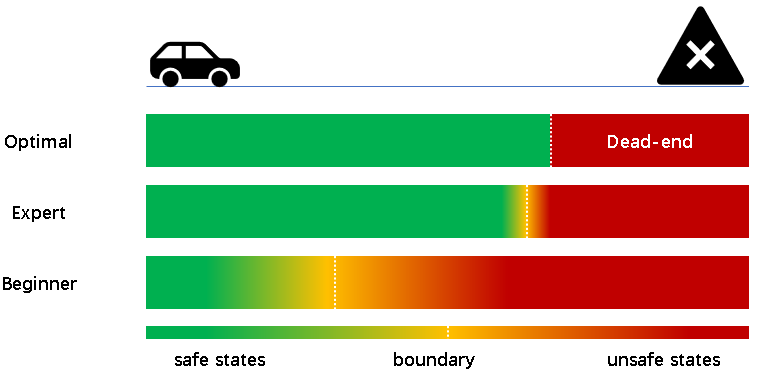}
    \caption{
        The agent needs to control the brake of the car to avoid a collision.
        A safety critic evaluates the safety of the policy in all states and combines it with a threshold value to obtain the boundary that divides the safe and unsafe states.
        At a given initial speed, no policy can avoid a collision when the car is close enough to an obstacle, and these states are called dead-ends.
        The boundary that will and only identifies all dead-ends as unsafe states is called the optimal boundary.
        Since the optimal safe policy is safe in all states except dead-ends, by evaluating the safety of the optimal safe policy, it is possible to distinguish whether a state is a dead-end or not.
        In contrast, suboptimal safe policies can lead the safety critic to conservatively consider more states as unsafe.
        % Therefore, finding the optimal boundary is closely related to the optimal safe policy and is equivalent to the discovery of dead-ends.
    }
    \label{fig_illustration}
\end{figure}

To learn a safe policy that satisfies state-wise safety constraint in ``safe critical" task, the agent needs to evaluate the safety of each state and avoid entering unsafe states \cite{zhao2023state}.
In some Safe RL algorithms, the agent's awareness of safety of a state is achieved by the safety critic that evaluates the safety of the task policy in states.
The safety critic and a safety threshold together construct a boundary that divides the state space into safe and unsafe subspaces, as shown in \Reffig{fig_illustration}.
% \COMMENT{1 policy-related is not clear. There are two types of polices, which one is the policy you mentioned?  2 you often claimed that a policy will consider or decide which state is safe or unsafe. Isn't it wrong
%     ? It is the safety critic who evaluate the safety of a state }
The division of the state space depends on the policy, and sub-optimal policies will lead to more states being considered as unsafe, thus limiting agent exploration.

% Previous experiments have shown that it is difficult to train a policy that has both high task performance and high safety.
% \cite{thananjeyan2021recovery} introduces a recovery policy and a behavior correction mechanism so that the task policy only needs to focus on the return.
% However, the estimation of safety in [8] is still based on task policy.
% This greatly limits exploration since it is prone to misjudge under-explored states as unsafe, which in turn leads to inadequate collection of trajectories to correct the safety critic.

In \cite{thananjeyan2021recovery}, a recovery policy and a behavior correction mechanism are introduced. The task policy and the recovery policy are trained simultaneously to improve task performance and to satisfy safety constraints respectively.
However, \cite{thananjeyan2021recovery} inappropriately consider the safety of states as the safety of the task policy, which leads to a conservative agent since the agent is prone to misjudge under-explored states as unsafe.
This greatly limits exploration, which in turn leads to inadequate collection of trajectories to correct the safety critic.
% As shown in \Reffig{fig_illustration}, estimating the safety of states using undertrained policies can easily mistake safe states for unsafe states, result in limitation of exploration.
% In the task shown in \Reffig{fig_illustration}, no policy can avoid a collision in some states, which are called dead-end, where the vehicle is close enough to the obstacle.

An example is given to illustrate the problem.
A beginner learns to drive.
Due to his poor driving skills, he ends up driving off the road several times on narrow roads, which led him to believe that driving on narrow roads is a very dangerous behavior and prevents him from continuing to explore in this environment.
To solve this problem, an expert who can drive the car back on the road before the imminent danger occurs should be introduced.
Thus, the beginner only needs to focus on the reward cumulating since the expert can take over before the imminent safety violation.
In this way, the poor skill of the beginner will no longer prevent risky exploration.

Inspired by this, we propose our safe RL framework \emph{Safe Reinforcement Learning with Dead Ends Avoidance and Recovery} (DEA-RRL), following the Recovery RL framework\cite{thananjeyan2021recovery} and decoupled RL framework\cite{schafer2021decoupling}.
A recovery policy and a safety critic are trained with the goal of maximizing safety.
% which act as a coach when training task policy.
When the task policy generates an action, the safety critic judges the safety of the action and decides whether to correct the action before interacting with the environment.
Compared with RRL\cite{thananjeyan2021recovery}, the safety critic in our approach measures the safety of the recovery policy rather than the task policy, which reduces the conservatism and relaxes the constraint on exploration without increasing the risk of violations.
% The difference between our approach and shielding-based approachs\cite{alshiekh2018safe,odriozola2023shielded}, which also include a shield mechanism  for identifying and correcting unsafe actions, is that the safety rules in our approach are learned from offline data and do not require manual design or prior knowledge of environmental dynamics.

% \COMMENT{learning the safety critic and policy in offline setting can rise other concerns from the reviewer. }

The advantages of our approach are as follows:
\begin{itemize}
    \item Our method allows task policy to fully explore the environment and significantly improve task performance in complex environments;
    \item We show theoretically that our approach can learn optimal boundary, which is equivalent to the discovery of dead-ends.
          % allows the agent to explore all states outside the dead-ends, which is the largest range that can be explored safely.
    \item Our method completely decouples the task policy and safe policy, so that the recovery policy can be plug-and-play within other task policies without fine-tuning.
          % \item Our method can be combined with offline RL algorithms to train in a completely offline setting to obtain safe policies.
\end{itemize}
% A series of continuous control experiments are designed to demonstrate that prior algorithms are too conservative in satisfying constraints to explore sufficiently that end up with lower rewarding.
% In contrast, our framework is able to balance exploration and exploitation and find a policy with high task performance and few violations of constraints.

\section[sec:related_work]{RELATED WORK}

% \subsection{Safe RL Using Prior Knowledge}
\subsection{Safe RL}

% \COMMENT{The review is not organized nor complete. Should the methods be classified into ``asymptotic safety'' after convergence and ``exact safety'' during training?}

% \COMMENT{should refer to this reference https://arxiv.org/pdf/2302.03122.pdf}

Safe RL addresses two main safety-related problems: asymptotic safety and in-training safety of policies.
Asymptotic safety means the safety of policies after convergence, which is commonly achieved by reward penalties and cost constraints.
Lagrange Relaxtion\cite{ha2020learning} is the most widely used method due to its simplicity, and other methods such as Trust Regions\cite{achiam2017constrained}, Lyapunov-based\cite{yang2020projection}, Guide Policy\cite{kim2022safety} are proposed for their stricter safety guarantees, fewer violations during the training and having faster convergence.

Although these approaches do find safe policies after training, they learn safety by trial-and-error as the same as traditional RL algorithms, which means violations are inevitable before convergence.
% that it is unacceptable in real word environment.
% Obviously, it is impossible to learn a safe policy and not violate safety constraints directly in an unknown environment without any priori knowledge since the agent cannot identify the safety of the action before interaction.
% Obviously, it is impossible to learn a safe policy and not violate safety constraints directly in an unknown environment without any priori knowledge.
% Other approaches assume a dynamics model of the environment or a large amount of trajectory data collected by other agents in this environment is existed\cite{zhao2023state}, from which a set of safety policies can be extracted.
To ensure in-training safety, state-wise safety constraints and prior knowledge of the environment are utilized \cite{zhao2023state}.
\cite{cheng2019end}\cite{shao2021reachability}\cite{wachi2020safe} assume a white-box or a black-box environment dynamics model is available, limiting policy optimizing in a safe policy set which is constructed based on the known model.
However, these assumptions are hard to meet in real-world problems.

% RRL \cite{thananjeyan2021recovery} directly train a task policy and a recovery policy from offline dataset, which maximizing the reward and correnting unsafe actions separatly, combined through control switching.

RRL \cite{thananjeyan2021recovery} trains a recovery policy using offline data to guarantee the safety of the online training process.
\cite{luo2021mesa} combines RRL and Meta-RL\cite{finn2017model}, improving the generalizability of RRL by enabling pre-trained safety critic and recovery policy to be quickly adapted to different tasks during the fine-tuning phase.
% The approach proposed in this paper belongs to the second category, utilising the priori information of environment by learning the safety evaluation function and recovery policy offline, and guaranteeing the safety during task policy exploration and learning during the online training phase with the safety evaluation function and safe policy.
Although these methods can achieve in-training safety, they are prone to obtain over-conservative polices due to the reason explained in \Refsec{sec:introduction}.

% \COMMENT{a conclusion is expected, \eg{most of these methods either make strong assumptions of the environmental dynamics or are overly conservative.}}

\subsection{Decoupling Performance and Safety}
Decoupling performance and safety offers new ideas for solving the exploration-exploitation dilemma\cite{schafer2021decoupling,whitney2021decoupled}, because separating processes of maximizing reward and guaranteeing safety prevents policies from under-performing due to over-conservatism.
\cite{srinivasan2020learning} and \cite{thananjeyan2021recovery} introduce an implicit and an explicit safe policy respectively.
% decoupling the original task into two parts.
% By decoupling, 
The task policy no longer needs to consider safety explicitly when being updated. %, but when interacting with the environment, the agent still needs to select an action such that the task policy satisfies the safety constraint after executing this action.
However, the identification of unsafe actions in these methods strongly correlated with the task policy, thus the agent is still prevented from obtaining optimal policies by exploration-exploitation dilemma.
% Therefore, we call it \emph{simi-decopuled}.
Although \cite{zhang2022safety} achieves full decoupling of exploration and exploitation, it focuses only on the rewards and ignores safety in training.
% We consider a complete decoupling of task policy and safety, with no need to consider the safety of the task policy when interacting with the environment, which having the advantage of extending the range of agents that can be explored.% , but also allows our recovery policy to secure any task policy, including those that have been trained.

\subsection{Dead-Ends Discovery and Avoidance}
The concept of dead-ends discovery (DeD) was introduced in \cite{fatemi2019dead} and later applied to the medical field\cite{fatemi2021medical}. %, \COMMENT{remove the following sentence?} where DeD model is trained in an offline dataset to determine whether a treatment is appropriate for a patient.
%\COMMENT{this citation can be removed}
\cite{killian2023risk} combines risk sensitivity with DeD model, which allows dead-ends to be identified earlier.
\cite{thomas2021safe} also proposes ``irrecoverable state" with a similar meaning to dead-ends state, preventing dangerous situations from occurring through reward shaping and model-based rollout\cite{janner2019trust}.

We argue that distinguishing dead-ends states from normal states is crucial for improving the performance of safe RL, as it identifies the broadest range of policies that can be explored safely.
The safety critic in RRL \cite{thananjeyan2021recovery} could not distinguish dead-ends states, as explained later in \Refsec{sec:method:rrl}.
%  because suboptimal task policies may still choose unsafe actions in non-dead-ends states, causing these states to be misidentified as dead-ends states eventually.

% This paper introduces the concept of dead-ends Avoidane in continuous control tasks and demonstrate that safety critic can be used to identify Dead-end states in our setup.

% \COMMENT{Offline RL is not related to safe, therefore I moved them to the offline training sections.}
% \subsection{Offline RL}
% In offline RL, task policy is trained in offline data directly\cite{lange2012batch}, which preventing violations occurring by skipping interaction with environment\cite{fujimoto2019off}.
% Due to distribution shifts and out-of-distribution actions\cite{kumar2019stabilizing}\cite{kumar2020conservative}\cite{kostrikov2021offline}, it is difficult for offline RL to learn a good policy directly from a dataset of average quality.
% In contrast, it is easier to learn environmental safety from offline data, as only enough violation trajectories are required, and the performance of behavioral strategies is not necessary.
% In the pre-training phase of our approach, safety critic also faces the problem of overestimation of safety due to out-of-distribution actions, which we tackle using implicit Q learning\cite{kostrikov2021offline}.
% Compared to RRL's single-step dynamic programming, our approach is multi-step dynamic programming\cite{kostrikov2021offline}, which allows for more secure trajectories to be planned by conbiming sub-optimal trajectories.

\section{Preliminary}
% \subsection{Markov Decision Processes}
% \COMMENT{remove this subsection and merge it to CMDP}
% We consider RL under Markov decision processes (MDPs) $\mathcal M=(\mathcal{S, A, R}, P, \gamma)$ \cite{altman1999constrained}, where $\mathcal S$ is the state space, $\mathcal A$ the action space, $\mathcal {R:S\times A}\rightarrow\mathbb{R}$ the reward function, $P:\mathcal{S\times A \times S}\rightarrow \mathbb [0,1]$ the transition function and $\gamma\in[0,1)$ the discount factor.
% Let $\Pi$ be the set of Markovian stationary policies.
% Given policy $\pi\in\Pi:\mathcal{S\rightarrow P(A)}$ maps states to action distributions and $\pi(a|s)$ denotes the probability of choosing action $a$ in state $s$.
% The objective of standard MDPs is to find a task policy $\pi_{task}\in\Pi$ to maximise the expected return $\mathcal J$:
% \begin{equation}
%     \label{object_function}
%     \mathcal J(\pi)=\mathbb{E}_{\tau\sim \pi,P}[\sum_{t=0}^\infty\gamma^t \mathcal R(s_t,a_t,s_{t+1})]
% \end{equation}

\subsection{Constraint Markov Decision Processes}
% Constraint Markov Decision Process (CMDPs) $\mathcal M=(\mathcal{S, A, R}, P, \gamma, \rho, \mathcal{C}, \gamma_{safe}, \epsilon_{safe})$ \cite{altman1999constrained} is defined based on  MDPs by adopting $\mathcal {C:S\times A}\rightarrow\mathbb{R}$ the cost function, $\gamma_{safe}\in[0,1)$ the discount factor for cost and $\epsilon_{safe}\in\mathcal R$ the safe threshold.
We consider safe RL under Constraint Markov Decision Process (CMDPs) $\mathcal M=(\mathcal{S, A, R}, P, \gamma, \rho, \mathcal{C}, \gamma_{safe}, \epsilon_{safe})$ \cite{altman1999constrained}, where $\mathcal S$ is the state space, $\mathcal A$ the action space, $\mathcal {R:S\times A}\rightarrow\mathbb{R}$ the reward function, $P:\mathcal{S\times A \times S}\rightarrow \mathbb [0,1]$ the transition function, $\gamma\in[0,1)$ the discount factor for reward, $\mathcal {C:S\times A}\rightarrow\mathbb{R}$ the cost function, $\gamma_{safe}\in[0,1)$ the discount factor for cost and $\epsilon_{safe}\in\mathcal R$ the safe threshold.
Let $\Pi$ be the set of Markovian stationary policies.
Given policy $\pi\in\Pi:\mathcal{S\rightarrow P(A)}$ maps states to action distributions and $\pi(a|s)$ denotes the probability of choosing action $a$ in state $s$.
The task performance of $\pi$ is defined as the discounted cumulative reward $\mathcal J(\pi)$:
\begin{equation}
    \label{object_function}
    \mathcal J(\pi)=\mathbb{E}_{\tau\sim \pi,P}[\sum_{t=0}^\infty\gamma^t \mathcal R(s_t,a_t,s_{t+1})]
\end{equation}

Similar to \cite{arulkumaran2017deep}, state-cost function $V_c$ and action-cost function $Q_c$ are introduced to indicate the expected cumulative cost of $\pi$ in $s$:
\begin{equation}
    \label{state_cost_function}
    V_c^{\pi}(s)=\mathbb E_{\tau\sim\pi,P}[\sum_{t=0}^{\infty}\gamma_{safe}^t\mathcal C(s_t,a_t,s_{t+1})|s_0=s]
\end{equation}
\begin{equation}
    \label{action_cost_function}
    Q_c^\pi(s,a)=\mathbb E_{s'\sim P(\cdot|s,a)}V_c^{\pi}(s')
\end{equation}

Under the state-wise safety constraint formulation \cite{zhao2023state}, we define the set of safe policies
\begin{equation}
    \Pi_c=\{\pi\in\Pi|\forall s\in\mathcal S, V_c^{\pi}(s)<\epsilon_{safe}\}
\end{equation}
The objective of CMDPs is to find a policy that maximizes \Refeq{object_function} in the set of safe policies $\Pi_c$:
\begin{equation}
    \pi_{task}^*=\max_\pi\mathcal J(\pi),\rm{s.t.}\pi\in\Pi_c
\end{equation}

\subsection{Safe Markov Decision Processes}

% CMDPs focus on the expectation of cumulative costs, making it allow some hazardous events with low costs to occur.
In safety-critical tasks, any unsafe action is fatal.
Therefore, we define SMDPs, a special case of CMDPs to describe this kind of tasks.
\begin{definition}
    Safe Markov Decision Processes (SMDPs) $\mathcal M=(\mathcal {S, A, R}, P, \gamma, \rho, \mathcal C, \gamma_{safe}, \epsilon_{safe})$, a special case of CMDPs where episodes terminate after any danger occurred.
    In SMDPs, the cost function is a binary indicator of the safety of the state:
    \begin{equation}
        \mathcal C(s_t,a_t,s_{t+1})=\left\{
        \begin{aligned}
             & 1,\ safety\ violation \\
             & 0,\ otherwise
        \end{aligned}
        \right.
    \end{equation}
\end{definition}
% \COMMENT{during training?}.
% \COMMENT{ Terminates IFF a danger occurs? no other terminal such as a goal is reached?}
% With the assumption made, to simplify the problem, the episode terminates when and only when a danger occurs.

Observe that if $\gamma_{safe}$ = 1, $Q_c^\pi$ indicates the probability of ending up with a failure state in the future with $\pi$.
$Q_{\phi, c}^\pi$, parameterized by $\phi$, can be optimized by minimizing the MSE loss:
\begin{equation}
    \label{obj_safe_critic}
    \begin{aligned}
         & \mathcal L_{Q_c}(\phi;\pi)=\frac{1}{2}(Q_{\phi,c}^{\pi}(s_t,a_t)-(c_t+(1-c_t)                     \\
         & \gamma_{safe}\mathbb E_{a_{t+1}\sim{\pi(\cdot|s_{t+1})}}Q_{\hat\phi,c}^{\pi}(s_{t+1},a_{t+1})))^2
    \end{aligned}
\end{equation}
where $(s_t,a_t,s_{t+1},c_t)$ is the transition sampled from offline data or replay buffer and $c_t$ is short for $\mathcal C(s_t, a_t, s_{t+1})$.
\Refeq{obj_safe_critic} can be considered to be the policy evaluation of $\pi$ in terms of safety.
% We classify the set of states $\mathcal S$ in SMDPs.
\begin{definition}
    \label{def1}
    The state space is divided into three subspaces:
    \begin{itemize}
        \item $\mathcal S_{fail}$: Failure state. Indicates the end of an episode due to a danger.
        \item $\mathcal S_{dead}$: Dead-ends state. Any dead-ends state will transform into a failure state, regardless of the policy $\pi$ the agent takes ($\pi\in\Pi$).
        \item $\mathcal S_{safe}$: Safe state. $\mathcal S_{safe}$=$\mathcal S\setminus(\mathcal S_{fail}\cup\mathcal S_{dead})$.
    \end{itemize}
    Therefore,
    \begin{equation}
        \label{eq_ssafe}
        \forall s\in\mathcal S_{safe}, \exists (a,s')\in(\mathcal {A, S}_{safe}), P(s, a, s')>0
    \end{equation}
\end{definition}

The cost function in SMDPs can also be expressed as
\begin{equation}
    \label{cost_function}
    \mathcal C(s_t,a_t,s_{t+1})=\mathbb I(s_{t+1}\in\mathcal S_{fail})
\end{equation}

Similar to \cite{thomas2021safe}, we assume that a safety violation must come fairly soon after entering any dead-ends states region:

\begin{assumption}
    \label{ass1}
    There exists a horizon $H\in\mathbb N$, that any trajectory starting from $s_0\in\mathcal S_{dead}$ will end up in $H$ steps.
\end{assumption}

% A safe agent should ensure that it always stays in $\mathcal S_{safe}$.
We additionally introduce a sampling policy for training the safety critic $\bar\pi$ and modify the definition of the set of safe policies as follows:
\begin{equation}
    \label{safe_policies_set}
    \begin{aligned}
         & \Pi_c^{\bar\pi}=\{\pi\in\Pi|\forall (s,a)\in(\mathcal S_{safe},\mathcal A)\ \rm{and}\ \pi(a|s)>0, \\
         & Q_c^{\bar\pi}(s,\pi(s))<\epsilon_{safe}\}
    \end{aligned}
\end{equation}
% In most approaches of Safe RL without decoupling, $\bar\pi$ is the same as $\pi_{task}$, including RRL\cite{thananjeyan2021recovery}.

Similar to CMDPs, the objective of SMDPs is to find a policy with \Refeq{object_function} in the set of safe policies $\Pi_c^{\bar\pi}$:
\begin{equation}
    \pi_{task}^*=\max_\pi\mathcal J(\pi),\rm{s.t.}\pi\in\Pi_c^{\bar\pi}\label{obj_smdps}
\end{equation}

% \COMMENT{should be placed here}

%  which is also named \emph{irrecoverable} in \cite{thomas2021safe}.
% So that we introduce a cost function $\mathcal {C:S\times A\times S}\rightarrow\{0,1\}$ in CMDPs to denote the cost incurred by performing an action in a given state.
% When the episode terminates, the agent will receive a cost equal to 1, indicating that the agent is using an unsafe action to interact with the environment; in all other cases, the agent receives a cost of 0.

\section{METHODS}
\subsection{Recovery RL}
\label{sec:method:rrl}
% \COMMENT{I moved the comments to specified method to here.}
In most approaches of Safe RL without decoupling, $\bar\pi$ is the same as $\pi_{task}$, including RRL\cite{thananjeyan2021recovery}.
In pretrain phase, RRL trains safety critic $Q_{\phi, c}^{\pi_{task}}$ and recovery policy $\pi_{\theta,rec}$ (parameterized by $\theta$) by minimizing $\mathcal L_{Q_c}(\phi;\pi_{task})$ and maximizing $\mathcal J_{\pi_{rec}}(\theta; \pi_{task})$ in \Refeq{obj_safe_critic} and \Refeq{obj_rec_p}.
\begin{equation}
    \mathcal J_{\pi_{rec}}(\theta;\bar\pi)=-\mathbb E_{s\sim \mathcal D}[Q_{c}^{\bar\pi}(s,\pi_{\theta,rec}(\cdot|s))]\label{obj_rec_p}
\end{equation}
In fine-tune phase, unsafe actions will be corrected by $\pi_{rec}$:
\begin{equation}
    \label{behavior_correction}
    a_t=\left\{
    \begin{aligned}
         & a^{\pi_{task}}, Q_{\phi, c}^{\bar\pi}(s_t, a^{\pi_{task}})<\epsilon_{safe} \\
         & a^{\pi_{rec}}, otherwise
    \end{aligned}
    \right.
\end{equation}
% where $\bar\pi$ is $\pi_{task}$ in RRL.
Similar to \Refeq{obj_smdps} The objective of RRL can be expressed as:
\begin{equation}
    \pi_{task}^*=\max_\pi\mathcal J(\pi),\rm{s.t.}\pi\in\Pi_c^{\pi}
\end{equation}

% \COMMENT{verify, pi task is disabled?}
Because $\pi_{task}$ has not been trained during pretrain phase that it is unable to avoid unsafe situations, resulting in an overestimation of $Q_{\phi,c}^{\pi_{task}}$  which represents a conservative estimate of safety.

\subsection{Dead-ends Discovery and Avoidance}
Different from RRL, our proposed DEA-RRL ensure safety by distinguishing between dead-ends states and safe states and only prevent the agent from entering dead-ends.
In the pretrain phase, DEA-RRL trains safety critic $Q_{\phi, c}^{\pi_{rec}}$ and recovery policy $\pi_{\theta,rec}$ by minimizing $\mathcal L_{Q_c}(\phi;\pi_{rec})$ and $\mathcal J_{\pi_{rec}}(\theta;\pi_{rec})$ as in \Refeq{obj_safe_critic} and \Refeq{obj_rec_p}, which is completely decoupled from $\pi_{task}$.
This is equivalent to solving the optimal Bellman equation\cite{arulkumaran2017deep} for safety.
Thus, the optimal recovery policy and the corresponding policy evaluation function can be obtained.
\begin{equation}
    \begin{aligned}
         & \pi_{rec}^*(\theta)=\min_\theta\mathbb E_{\tau\sim P,\pi_{rec}(\theta)}[Q_{c}^{\pi_{rec}}(s,\pi_{rec}(s;\theta))] \\
         & =\min_\theta\mathbb E_{\tau\sim P,\pi_{rec}(\theta)}[\sum_{t=0}^{\infty}\gamma_{safe}^tc_t]
    \end{aligned}
\end{equation}
By using the same behavior correction mechanism as \Refeq{behavior_correction} where $\bar\pi$ is $\pi_{rec}^*$, the object of DEA-RRL can be expressed as:
% \COMMENT{verify the equation $\pi\in\Pi_c^{\pi_{rec}^*}$?}
\begin{equation}
    \pi_{task}^*=\max_\pi\mathcal J(\pi),\rm{s.t.}\pi\in\Pi_c^{\pi_{rec}^*}
\end{equation}

\subsection[sec:proof]{Theoretical Proof}
We will illustrate the advantages of DEA-RRL over RRL theoretically.
%  level in this section.
\begin{theorem}
    \label{optimal_safety_critic}
    $\Pi_c^{\pi_{task}}$ and $\Pi_c^{\pi_{rec}^*}$ are the accessible space for $\pi$ to explore safely in RRL and DEA-RRL respectively, we have $\Pi_c^{\pi_{task}}\subseteq\Pi_c^{\pi_{rec}^*}$.
\end{theorem}
\emph{Proof}: Since $Q_{c}^*$ (the shorthand of $Q_{c}^{\pi_{rec}^*}$) is the optimal cost value function, we have $Q_{c}^*(s,a)\leq Q_{c}^{\pi_{task}}(s,a)$ for every $(s,a)$.
As in \Refeq{safe_policies_set}, $\forall(\pi,s,a)\in(\Pi_c^{\pi_{task}},\mathcal S_{safe},\mathcal A)$ and $\pi(a|s)>0$,
$$Q_c^*(s,\pi(s))\leq Q_c^{\pi_{task}}(s,\pi(s))<\epsilon_{safe}$$
therefore $\pi\in\Pi_c^{\pi_{rec}^*}$, which means $\Pi_c^{\pi_{task}}\subseteq\Pi_c^{\pi_{rec}^*}$.
% Therefore, DEA-RRL provides larger accessible space for exploration than RRL, reducing conservatism without loss of safety.

We will show that with appropriate $\epsilon_{safe}$, $Q_{c}^*$ can be used to identify dead-ends states.% \COMMENT{unify the terms, dead-ends or non-recoverable} non-recoverable states.
%  which indicating the maximum field that can be safely explored.

\begin{lemma}
    \label{ded}
    Suppose that Assumption \ref{ass1} holds and uncertainties are ignored in the environment \ie{$P:\mathcal{S\times A\times S}\rightarrow\{0, 1\}$},
    \begin{itemize}
        \item $\forall (s,\pi)\in(\mathcal S_{fail},\Pi),\ V_c^{\pi}(s)=V_c^*(s)=1,$
        \item $\forall (s,\pi)\in(\mathcal S_{dead},\Pi),\ V_c^{\pi}(s)\geq V_c^*(s)\geq\gamma_{safe}^{H-1}$
        \item $\forall (s,\pi)\in(\mathcal S_{safe},\Pi),\ V_c^{\pi}(s)\geq V_c^*(s)=0$
    \end{itemize}
\end{lemma}
\emph{Proof}:
Since $V_c^*$ is the optimal value function, we have $V_c^{\pi}(s)\geq V_c^*(s)$ for all $s\in\mathcal S$.
By the definition of cost function \Refeq{cost_function} and state-cost function \Refeq{state_cost_function}, $\forall s\in\mathcal S_{fail}, V_c^{\pi}(s)=1$.
Assumption \ref{ass1} shows that episode would terminate after at most $H$ steps since the agent reach dead-ends state, we can derive from \Refeq{state_cost_function} that
\begin{equation}
    \begin{aligned}
         & V_c^{\pi}(s)=\mathbb E_{\tau\sim\pi,P}[\sum_{t=0}^{\infty}\gamma_{safe}^t\mathcal C(s_t,a_t,s_{t+1})|s_0=s\in\mathcal S_{dead}] \\
         & \geq\sum_{t=0}^{H-2}\gamma_{safe}^t\ast 0 + \gamma_{safe}^{H-1}\ast 1=\gamma_{safe}^{H-1}.
    \end{aligned}
\end{equation}
By \Refeq{eq_ssafe} in Definition \ref{def1}, for every $s\in\mathcal S_{safe}$, there always exists at least one action $a$ such that $s'\sim P(\cdot|s, a), s'\in\mathcal S_{safe}$.
Start at $s\in\mathcal S_{safe}$, agent would never reach dead-ends state by choosing the safe action, which means that $V_c^*(s)=0$.

% \COMMENT{Corollary? not lemma}
\begin{theorem}
    \label{dea}
    Suppose that Assumption \ref{ass1} holds and uncertainties are ignored in the environment, and let
    \begin{equation}
        \epsilon_{safe}=\gamma_{safe}^{H}
    \end{equation}
    Then, with behavior correction mechanism showed by \Refeq{behavior_correction}, the agent will be prevented from reaching dead-ends states.
\end{theorem}
\emph{Proof}:
According to \Refeq{behavior_correction}, the actions allowed to be performed satisfy:
\begin{equation}
    Q_c^\pi(s,a)=\mathbb E_{s'\sim P(\cdot|s,a)}\gamma_{safe}V_c^{\pi}(s')<\gamma_{safe}^{H}
\end{equation}
therefore
\begin{equation}
    V_c^{\pi}(s')<\gamma_{safe}^{H-1}, \forall s\in\mathcal S_{safe}
\end{equation}
which ensuring $s'\in\mathcal S_{safe}$.
Also because of the initial state $s_0\in\mathcal S_{safe}$, it is ensured that the agent is always explored in safe states.

Theorem \ref{optimal_safety_critic} shows that DEA-RRL provides larger accessible space for exploration than RRL, reducing conservatism without loss of safety.
Lemma \ref{ded} and theorem \ref{dea} indicate that  starting from safe states, both RRL and DEA-RRL are able to prevent agent entering dead-ends states with behavior correction mechanism and appropriate $\epsilon_{safe}$.
Notice that in RRL, it is not ensured that $V_c^{\pi_{task}}(s)<\gamma_{safe}^{H-1}, \forall s\in\mathcal S_{safe}$, thus RRL will recognize some safe states as dead-ends states due to its over-conservatism on under-explored states.

\subsection{Offline Pretrain}
% The concepts of single-step dynamic programming (SSDP) and multi-steps dynamic programming (MSDP) are introduced in \cite{kostrikov2021offline}, which can be used to explain the difference of pretrain phase between DEA-RRL and RRL.
The safety critic in RRL is entirely determined by $\pi_{task}$, thus $\pi_{rec}$ is equivalent to a single-step greedy policy that chooses the action with the smallest $Q_c^{\pi_{task}}$ in a state.
This is referred to as single-step dynamic programming (SSDP)\cite{kostrikov2021offline}.
By contrast, safety critic in DEA-RRL is determined by $\pi_{rec}$, giving $\pi_{rec}$ the ability to combine multiple sub-optimal trajectories into one optimal trajectory, hence it is referred to as multi-steps dynamic programming (MSDP)\cite{kostrikov2021offline}.

The bias in MSDP due to querying the $Q_c^{\pi}$ value of out-of-distribution (OOD) actions accumulates over the DP process, making MSDP more sensitive to OOD actions compared with SSDP.
In the training of safety critic, we prevent safety critic from overestimating the safety of an action by avoiding querying the $Q_c$ value of OOD actions.
While in the training of recovery policy, we can not completely ignore OOD actions especially in states where the offline data do not contain safe actions, for it is better to try an unknown and possibly safe action than to choose a known but certainly dangerous action.

% In the offline training phase, \COMMENT{is this significant? and why?} DEA-RRL uses multi-step dynamic planning while RRL uses single-step dynamic planning, therefore, DEA-RRL will face more severe overestimation caused by out-of-distribution (OOD) actions.
Inspired by implicit Q-learning (IQL) algorithm \cite{kostrikov2021offline}, we use Expectile Regression to train $Q_c^{\pi_{rec}}$, avoiding the impact of OOD actions.
Different from IQL, we use Advantage Policy Gradient instead of Advantage Weighted Regression used in \cite{kostrikov2021offline} to train $\pi_{rec}$, allowing an OOD action to be attempted in a state where all known actions are unsafe.

The state-cost function $V_c^{\pi}$ and action-cost function $Q_c^{\pi}$ is updated by minimizing following loss functions:
\begin{equation}
    \label{v_loss}
    \mathcal L_{V_c^{\pi}}(\psi)=\mathbb E_{(s,a)\sim\mathcal D}[L_2^{\tau}(Q_{\phi,c}^{\pi}(s,a)-V_{\psi,c}^{\pi}(s))]
\end{equation}
\begin{equation}
    \label{q_loss}
    \begin{aligned}
         & \mathcal L_{Q_c^{\pi}}(\phi)=\mathbb E_{(s,a,s')\sim\mathcal D}[(c(s,a,s')+ \\
         & (1-c(s,a,s'))\gamma_{safe}V_{\psi,c}^{\pi}(s')-Q_{\phi,c}^{\pi}(s,a))^2]
    \end{aligned}
\end{equation}
where $L_2^{\tau}$ is the expectile regression loss and $\mathcal D$ the offline dataset.
% In practice, we use Advantage Policy Gradient instead of Advantage Weighted Regression used in \cite{kostrikov2021offline}, because we cannot completely ignore OOD actions, especially in states where the offline data do not contain safe actions.

As illustrated in Algorithm \ref{alg_DEA-RRL-PT}, we provide the agent with $\mathcal D$ that contain several transitions from both safe trajectories and unsafe trajectories.
Safety critic and recovery are trained by minimizing corresponding objective functions without task policy anymore.
\begin{algorithm}
    \caption{DEA-RRL Pretrain Offline}
    \label{alg_DEA-RRL-PT}
    \begin{algorithmic}[1]
        \State Input: offline dataset $\mathcal D$, pretraining steps N
        \For{$steps, \leftarrow 1, N$}
        \State Sample a mini-batch $(s_t, a_t, s_{t+1}, c_t)$ from $\mathcal D$
        \State Update $\phi$ by minimizing  $\mathcal L_{Q_c^{\pi}}(\phi)$ (\Refeq{q_loss})
        \State Update $\psi$ by minimizing  $\mathcal L_{V_c^{\pi}}(\psi)$ (\Refeq{v_loss})
        \State Update $\theta$ by maximizing $\mathcal J_{\pi_{rec}}(\theta)$ (\Refeq{obj_rec_p})
        \EndFor
        % \State Output: safety critic $Q_c^{\pi_{rec}}$, recovery policy $\pi_{rec}$
    \end{algorithmic}
\end{algorithm}

\subsection{Online Training}
Any of the RL algorithms can be used to train $\pi_{task}$.
In our work, we utilize Soft Actor Critic algorithm (SAC) \cite{haarnoja2018soft}.
The process of online fine-tuning is illustrated in Algorithm \ref{alg_DEA-RRL-FT}, and we give some remarks on online training.
\begin{itemize}
    \item We relabel all actions with the action proposed by $\pi_{task}$ as the same as \cite{thananjeyan2021recovery}, which is important to achieve decoupled safe reinforcement learning, as it prompts the agent to view behavior correction as part of the environment.
    \item Since $\pi_{rec}$ ignores task performance, unsafe actions result in low reward by behavior correction, which enables $\pi_{task}$ to learn to avoid unsafe actions with reward feedback only.
    \item $\pi_{rec}$ is a Gaussian policy, we use the action mean rather than sampling over the distribution during behavior correction for safety.
    \item We choose not to fine-tune $Q_c^{\pi_{rec}}$ and $\pi_{rec}$ in online training, because a fixed behavior correction strategy provides the agent a stable MDP dynamic model, thus improving the stability of training.
    \item Finally, we use a small $\epsilon_{safe}$ to improve the safety of the algorithm due to aleatoric uncertainty and epistemic uncertainty,
\end{itemize}

\begin{algorithm}
    \caption{DEA-RRL Training Online}
    \label{alg_DEA-RRL-FT}
    \begin{algorithmic}[1]
        \State Input: safety critic $Q_c^{\pi_{rec}}$, recovery policy $\pi_{rec}$, task horizon $H$, training steps $N$
        \State Initialize replay buffer $\mathcal D_{task}\leftarrow\varnothing $
        \State $s_0\leftarrow env.reset()$
        \For{$steps, \leftarrow 1, N$}
        \For{$t\in\{1,...,H\}$}
        \If{$c_{t-1}=1\:or\:t=H$}
        \State $s_t\leftarrow env.reset()$
        \EndIf
        \State Sample $a^{\pi_{task}}$, $a^{\pi_{rec}}$ from $\pi_{task}$, $\pi_{rec}$
        \If{$Q_c^{\pi_{rec}}(s_t, a^{\pi_{task}})<\epsilon_{safe}$}
        \State $a_t=a^{\pi_{task}}$
        \Else
        \State $a_t=a^{\pi_{rec}}$ (behavior correction)
        \EndIf
        \State Execute $a_t$, Observe $s_{t+1}, r_t, c_t$
        \State $\mathcal D_{task}\leftarrow\mathcal D_{task}\cup(s_t, a^{\pi_{task}}, s_{t+1}, r_t)$
        \State Train $\pi_{task}$ on $\mathcal D_{task}$ by maximizing $\mathcal J(\pi)$
        \EndFor
        \EndFor
        % \State Output: $\pi_{task}$
    \end{algorithmic}
\end{algorithm}

\section{EXPERIMENTS}
In the experiments, we investigate whether our approach can:
\begin{itemize}
    \item exceed state-of-the-art algorithms in terms of task performance, post-training safety and in-training safety;
    \item improve the safety of the training process with minimal impact on task performance;
          % \item be combined directly with other trained algorithms to improve the safety of these algorithms in testing.
    \item obtain a task-independent behavior correction strategy that can ensure safety of other trained policies in testing.
\end{itemize}

\subsection{Domains}
\begin{figure*}[!ht]
    \centering
    \includegraphics[width=1\linewidth]{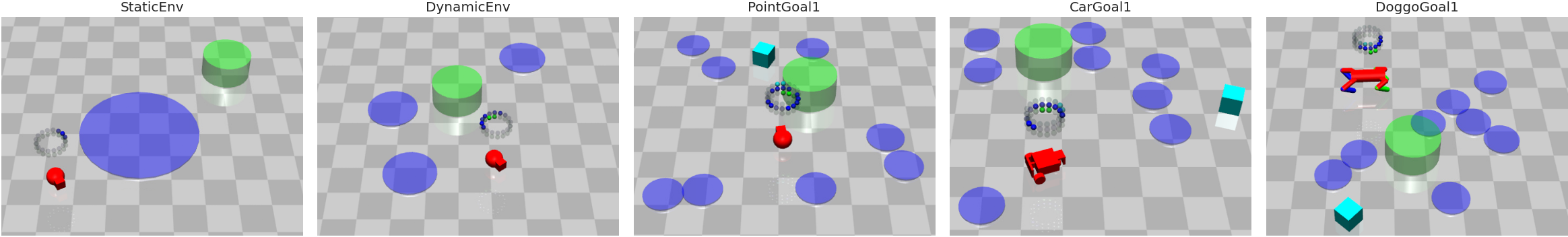}
    \caption{
        The experiments in this paper were conducted in the safety gym standard test environment.
        The figure shows five task environments from left to right, including StaticEnv, DynamicEnv, PointGoal1, CarGoal1 and DoggoGoal1.
        These environments are based on Safety Gym, where the task is to navigate a Point robot from its initial position to the target area (green area) and avoid entering unsafe areas (blue areas) during the process.
        PointGoal1, CarGoal1 and DoggoGoal1 can be seen as upgraded versions of DynamicEnv, which include more unsafe areas, larger action space and more complex robots.
    }
    \label{fig_envs}
\end{figure*}

Experiments were conducted under the standard safety reinforcement learning test environment Safety Gym\cite{ray2019benchmarking}.
As shown in \Reffig{fig_envs}, We selected the two simple environments (StaticEnv, DynamicEnv) set up in \cite{yang2021wcsac} and three more complex environments (PointGoal1, CarGoal1, DoggoGoal1) pre-defined in Safety Gym.
Unlike the general Safety Gym setup, we require that any violation of safety constraints will result in immediate termination of the episode, which corresponds to real world tasks.
% In realistic environmental tasks, agents should not be allowed to interact with the environment after violating safety constraints, or there is a high risk of further damage to the environment and other agents.
% This setup will encourage the agent to avoid entering any unsafe states \COMMENT{can RRL be benefiting from this setting?}, but it also significantly increases the difficulty of the task and poses an extreme challenge to the agent's capacity to balance exploration and exploitation.
This setup significantly increases the difficulty of the task and poses an extreme challenge to the agent's capacity to balance exploration and exploitation.

\subsection{Offline Data Collection}
Unlike general offline RL where the training results are influenced by the performance of the behavior policy that used for data sampling, Recovery RL framework requires offline data to contain a wide coverage of unsafe trajectories, allowing $Q_c$ to learn to identify unsafe actions efficiently.
The offline dataset we used contains 2M transitions including: (1) 1M transitions sampled from the replay buffer of SAC training, and (2) 1M transitions obtained by interacting with the environment using random actions.
The latter is included because in our experiments we find that the failed transitions in the SAC replay buffer tend to share similar characteristics.
The random sampling data can diversify the failed transitions in the offline data.
Inspired by \cite{feng2023dense}, we filter out the parts of the data that are less relevant to safety by keeping only 100 transitions before violations.

\subsection{Evaluation Metric}
The average cumulative reward (ACR) over episodes is used as a measure of the task performance, and a higher return indicates a better task performance.
The default reward functions provided by Safety Gym are used in our experiments, which define algorithm-independent tasks.
We use the average rate of constraint violation (AVR) in testing as a measure of asymptotic safety and the number of constraint violations (TV) in training as a measure that indicating in-training safety.
To highlight the differences between our method and RRL, we use the ratio of steps that behavior correction mechanism is used (ARR) as a measure of the extent to which the behavior correction mechanism intervene in training and testing phase.

\subsection{Comparisons with Baselines}
\begin{itemize}
    \item Unconstrained Baseline\cite{haarnoja2018soft}: We use SAC as a baseline for unconstrained methods, optimizing task performance and ignoring safety constraints.
    \item Worst-Case Soft Actor Critic (WCSAC)\cite{yang2021wcsac}: A combined Lagrangian relaxation and risk-sensitive approach that maximizing
          $$
              \mathcal J(\pi)-\lambda(\mathbb E_{\tau\sim P,\pi}[CVaR_{\alpha}(Q_c^{\pi_{task}}(s,a))]-\epsilon_{safe})
          $$
          where $CVaR_{\alpha}\doteq\mathbb E_{p^{\pi}}[Q_c^{\pi}|Q_c^{\pi}\geq F_C^{-1}(1-\alpha)]$ and $F_C$ is the CDF of $p^{\pi}(Q_c^{\pi}|s,a)$, updating policy parameters and $\lambda$ via dual gradient descent.
    \item RRL\cite{thananjeyan2021recovery}: A \emph{semi-decoupled} approach that preventing agent using unsafe action for which $Q_c^{\pi_{task}}(s,a)\geq\epsilon_{safe}$ by behavior correction mechanism.
    \item Implicit Q-Learning (IQL)\cite{kostrikov2021offline}: An offline RL algorithm that optimizing critic and actor by expectile regression and advantage weighted regression.
\end{itemize}
To better compare the ability of each method to balance safety and task performance, we let each method has similar asymptotic safety achieving by using different $\epsilon_{safe}$.

\begin{figure*}[!ht]
    \centering
    \includegraphics[width=1\linewidth]{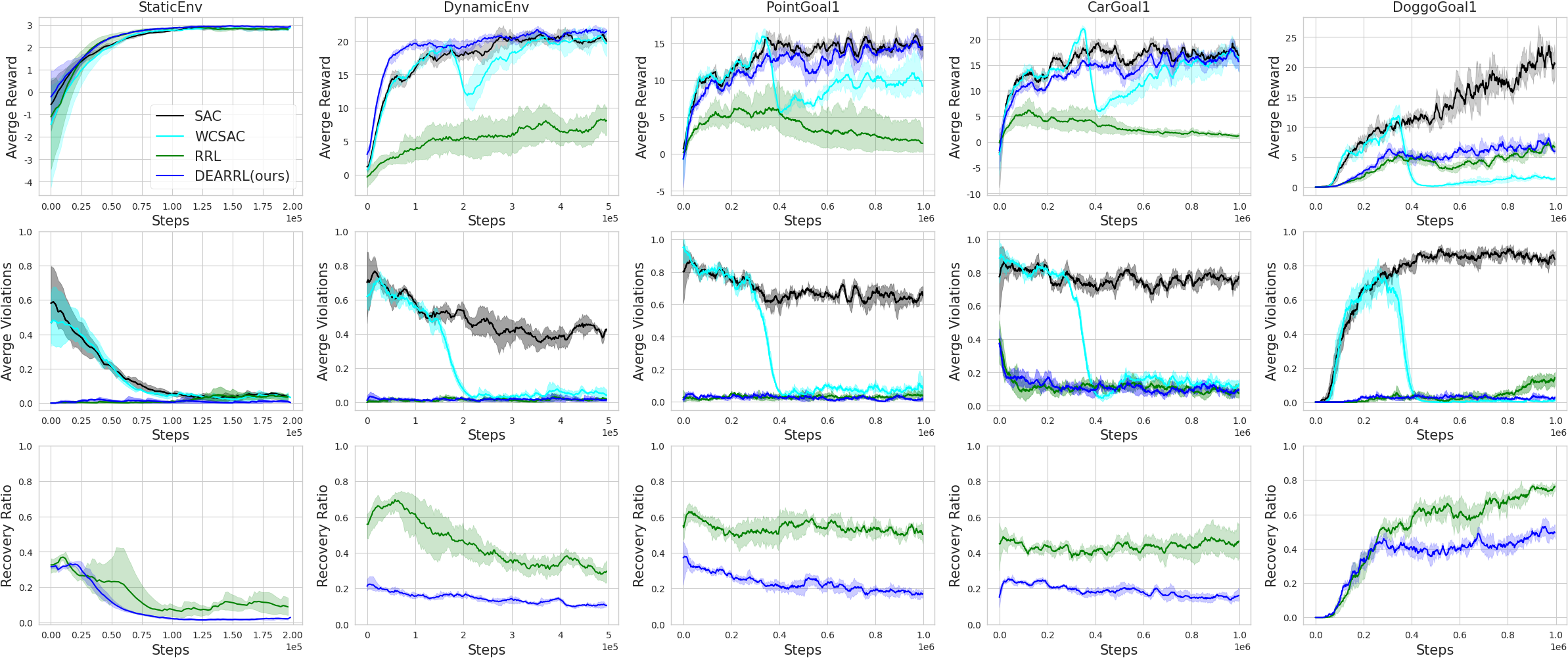}
    \caption{
        Training Curve. Means (solid lines) and variances (shaded) of the training curves for the four algorithms under different tasks, with the first row showing the average cumulative reward, the second row showing the average proportion of constraint violations over update steps and the third row showing the average use of behavior correction.
        We adapted $\epsilon_{safe}$ for RRL and DEA-RRL to make the two algorithms have similar safety in training. For each method we used random seeds for training.
    }
    \label{fig_main}
\end{figure*}

\begin{table*}[!ht]
    \caption{}
    \label{tab_main}
    \centering
    \renewcommand\arraystretch{1}
    \begin{threeparttable}
        \begin{tabular}{c|ccc|ccc|ccc|ccc|cc}
            \toprule
            \multicolumn{1}{c|}{\multirow{2}{*}{Environments}} & \multicolumn{3}{c|}{SAC} & \multicolumn{3}{c|}{WCSAC} & \multicolumn{3}{c|}{RRL} & \multicolumn{3}{c|}{DEA-RRL(ours)} & \multicolumn{2}{c}{IQL}                                                                                                                         \\
                                                               & ACR                      & AVR                        & TV                       & ACR                                & AVR                     & TV    & ACR   & AVR        & TV           & ACR             & AVR            & TV           & ACR             & AVR   \\
            \midrule
            StaticEnv                                          & 2.754                    & 0.043                      & 1642                     & \textbf{2.936}                     & \textbf{0.008}          & 1453  & 2.779 & 0.038      & 311          & 2.882           & 0.023          & \textbf{119} & 2.630           & 0.095 \\
            DynamicEnv                                         & 17.434                   & 0.519                      & 3330                     & 19.004                             & 0.069                   & 1451  & 7.479 & \textbf{0} & 92           & \textbf{20.784} & 0.018          & \textbf{86}  & 18.044          & 0.565 \\
            PointGoal1                                         & 12.609                   & 0.753                      & 12375                    & 7.811                              & 0.055                   & 5739  & 1.155 & 0.078      & 408          & \textbf{14.079} & \textbf{0.035} & \textbf{236} & 12.740          & 0.76  \\
            CarGoal1                                           & 16.727                   & 0.73                       & 14850                    & \textbf{18.672}                    & \textbf{0.035}          & 6199  & 1.827 & 0.043      & 1006         & 15.430          & 0.091          & \textbf{786} & 11.357          & 0.94  \\
            DoggoGoal1                                         & 16.956                   & 0.908                      & 31450                    & 1.055                              & \textbf{0.013}          & 10825 & 6.969 & 0.22       & \textbf{546} & 7.002           & 0.048          & 706          & \textbf{18.889} & 0.838 \\
            \bottomrule
        \end{tabular}
        The table records the average cumulative reward (ACR) and the average violation rate (AVR) for the policy trained by, and the number of violations during training (TV) for the SAC, WCSAC, RRL, DEA-RRL and IQL.
        Since IQL is trained completely offline, the number of violations during training is 0.
        The ACR and AVR are obtained by testing $\pi_{task}$ on 500 random episodes on average.
    \end{threeparttable}

\end{table*}

\begin{table*}[!ht]
    \caption{}
    \label{tab_decoupled}
    \centering
    \renewcommand\arraystretch{1}
    \begin{threeparttable}
        \begin{tabular}{c|ccc|ccc|ccc|ccc}
            \toprule
            \multicolumn{1}{c|}{\multirow{2}{*}{Environments}} & \multicolumn{3}{c|}{SAC(with RRL)} & \multicolumn{3}{c|}{IQL(with RRL)} & \multicolumn{3}{c|}{SAC(with DEA-RRL)} & \multicolumn{3}{c}{IQL(with DEA-RRL)}                                                                                     \\
                                                               & ACR                                & AVR                                & ARR                                    & ACR                                   & AVR   & ARR   & ACR             & AVR   & ARR   & ACR             & AVR   & ARR   \\
            \midrule
            StaticEnv                                          & 2.379                              & 0                                  & 0.417                                  & 1.308                                 & 0.005 & 0.426 & 2.673           & 0     & 0.009 & \textbf{2.947}  & 0     & 0.031 \\
            DynamicEnv                                         & 1.928                              & 0.004                              & 0.671                                  & 1.488                                 & 0.008 & 0.764 & \textbf{18.769} & 0.032 & 0.242 & \textbf{19.425} & 0.078 & 0.235 \\
            PointGoal1                                         & 5.462                              & 0.116                              & 0.546                                  & 3.669                                 & 0.064 & 0.581 & \textbf{10.596} & 0     & 0.275 & \textbf{9.926}  & 0.04  & 0.322 \\
            CarGoal1                                           & 0.093                              & 0.034                              & 0.614                                  & 0.144                                 & 0.094 & 0.602 & 8.165           & 0.028 & 0.334 & \textbf{10.783} & 0.116 & 0.297 \\
            DoggoGoal1                                         & 0.751                              & 0                                  & 0.987                                  & 0.656                                 & 0     & 0.941 & \textbf{13.285} & 0.002 & 0.780 & 9.407           & 0     & 0.846 \\
            \bottomrule
        \end{tabular}
        We compare the average cumulative reward (ACR) and average violation rate (AVR) of the algorithms by directly combining a pre-trained behavior correction mechanism with $\pi_{task}$ trained using SAC and IQL, where the parameters of RRL and DEA-RRL and $\epsilon_{safe}$ are set as the same as in the previous experiment.
        All data are derived from the average of 500 random episodes of test results.
    \end{threeparttable}

\end{table*}

\subsection{Results}

% \COMMENT{this paragraph can also be removed.}
% In the first part of the experiment, we compare the task performance and safety of different algorithms.
% In the second part we perform an ablative study on the impact of different $\epsilon_{safe}$ on task performance and safety of the algorithms.
% In the third part we show that, benefiting from a fully decoupling framework, our pre-trained behavior correction strategy can be directly combined with trained algorithms to improve safety of these algorithms in testing.

\subsubsection{Main Results}
The performance and safety of all methods are showed in \Reffig{fig_main} and \Reftab{tab_main}.
The results show that our method has high asymptotic safety and in-training safety in all tasks and similar cumulative reward to the unconstrained method in most tasks, suggesting that our method can substantially improve the safety of the algorithm with less impact on performance.

% In contrast, while the unconstrained approach achieves the highest returns for all tasks, both in training and after training the algorithm has a very high rate of constraint violations, which is unacceptable for real-world tasks.
WCSAC obtains a policy with higher safety than SAC after training for all tasks, however, since WCSAC can only learn information related to the safety of the environment during training, a large number of constraint violations are inevitable in the early stages of training.
% At the same time, because WCSAC uses the same policy to maximise performance and safety, it can face a serious exploration-exploitation dilemma, leading it to learn only sub-optimal policies with high safety and low returns in complex tasks.

IQL avoids interaction with the environment by training completely offline so that safety constraints are not violated during training, but a safe policy cannot be obtained after training.

RRL learns information about the safety of the environment in advance through pre-training and therefore maintains a low number of constraint violations throughout the training process.
% Although RRL performs well in simple environments, in complex environments RRL is limited by the simi-decoupling framework and still faces exploration exploitation dilemmas, making the algorithm less rewarding.
% Benefiting from the fully decoupled framework and the equivalence of $Q_c^{\pi_{rec}}$ in DEA-RRL for identifying Dead-ends states, $\pi_{task}$ in DEA-RRL allows full exploration of the environment without regard to safety, and the exploration space is not constrained by conservative safety estimates.
As shown in \Reffig{fig_main}, the ratio of steps RRL using behavior correction during training process is much higher than that of DEA-RRL, indicating a large amount of intervention in the training process of $\pi_{task}$, which is an important reason why RRL can only learn sub-optimal policies in complex environments.
The training curve of RRL has large variance, which is caused by the fact that $\pi_{task}$ is training in an unstable environment.
In contrast, DEA-RRL does not fine-tune $\pi_{task}$ and $Q_c^{\pi_{rec}}$ during online training, providing a stable MDPs for training $\pi_{task}$.

% \COMMENT{this explanation should be placed in the main results.}
We also remark that DEA-RRL has lower return than SAC in DoggoGoal1 because the environment are more difficult to explore with the introduction of the behavior correction mechanism, and it takes longer for agents to learn the optimal policy.

% Moreover, although DEA-RRL achieved lower return than SAC in DoggoGoal1, this was not due to DEA-RRL be too conservative, which is explained in \Refsec{sec_exp_decoupled}.
% \COMMENT{DoogoGoal1 should be explained here}

\subsubsection{Ablations}

We design ablation experiments to investigate the effect of settings in pre-training and online training on safety and task performance.
\begin{figure}[!ht]
    \centering
    \includegraphics[width=1\linewidth]{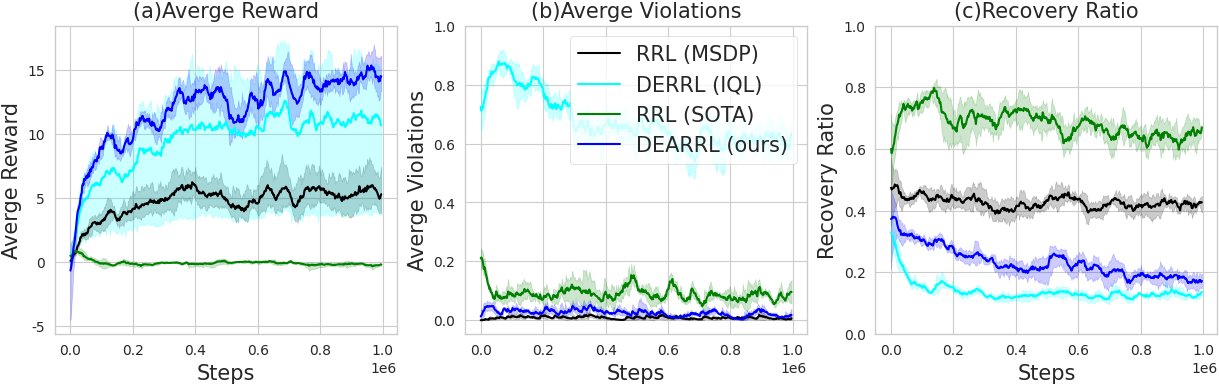}
    \caption{
        Ablations of offline pretraining.
        Average reward, average proportion of constraint violations and average proportion of use of behavioral correction for different methods.
        For each method we used same $\epsilon_{safe}=0.7$ and random seeds for training.
    }
    \label{fig_ablation1}
\end{figure}

We first compare the effects of using different training methods in offline training on the experimental results which are shown in \Reffig{fig_ablation1}.
RRL (MSDP) is a multi-step dynamic programming version of RRL, significantly reducing conservatism of RRL.
However, RRL (MSDP) is still highly conservative due to the OOD actions and has high recovery ratio and low cumulative reward.
DEARRL (IQL) extract a policy from safety critic using advantage weighted regression.
Since DEARRL (IQL) will only select known actions, it cannot guarantee safety in states where the dataset does not contain safe actions, resulting in an unsafe policy.

\begin{figure}[!ht]
    \centering
    \includegraphics[width=1\linewidth]{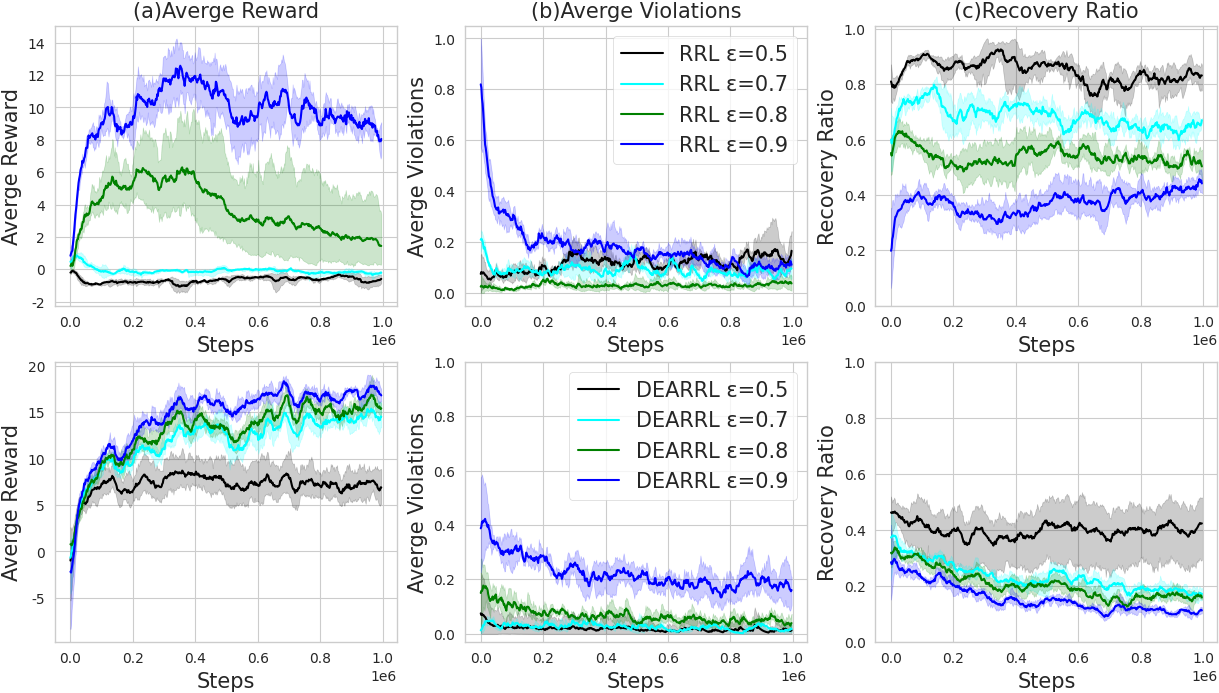}
    \caption{Ablations of online training. Average reward, average proportion of constraint violations and average proportion of use of behavioral correction for different $\epsilon_{safe}$ choices for RRL and DEA-RRL in PointGoal1. For each method we used random seeds for training.
    }
    \label{fig_ablation2}
\end{figure}

% Although setting $\epsilon_{safe}=\gamma_{safe}^H$ theoretically ensures that the agent remains in safe states, due to aleatoric uncertainty and epistemic uncertainty, we need to reduce $\epsilon_{safe}$ to improve the safety and stability of the algorithm in online training.
% In ablation experiments, 
We then compared the performance and safety of RRL and DEA-RRL under different $\epsilon_{safe}$.
As \Reffig{fig_ablation2} shows, as $\epsilon_{safe}$ decreases, RRL and DEA-RRL have the same performance, \ie{the average reward and the proportion of constraint violations decrease.}
Notice that with equal $\epsilon_{safe}$, DEA-RRL has a similar safety and far superior task performance than RRL.
DEA-RRL substantially reduces the influence of the behavior correction mechanism on $\pi_{task}$, and eventually enables algorithm to find a better $\pi_{task}$.

\subsubsection{Decoupled Framework}
\label{sec_exp_decoupled}
In DEA-RRL, $\pi_{rec}$ and $Q_c^{\pi_{rec}}$ are completely decoupled from $\pi_{task}$, so that the pre-trained $\pi_{rec}$ and $Q_c^{\pi_{rec}}$ can be directly combined with other $\pi_{task}$ trained to improve the safety of these algorithms in testing.
We combine the behavior correction strategy obtained by RRL and DEA-RRL pre-training with the other policies obtained through SAC and IQL training and test them on 500 random episodes, the results of which are shown in \Reftab{tab_decoupled}.
The results show that DEA-RRL can be combined directly with any RL algorithm to substantially improve the safety of these algorithms in testing at the expense of slightly lower return.
Notice that in some environment s, $\pi_{task}$ and $\pi_{rec}$ obtained by IQL (with DEA-RRL) even achieve the performance and safety of the online training method when trained completely offline.

Finally, we find that although the average episode length in testing increases significantly with the introduction of the behavior correction mechanism compared to using SAC and IQL only, there is only a slight improvement in the return.
The reason is that the introduction of the behavior correction mechanism actually changes the MDPs model so that policies trained using SAC and IQL that maximize the returns of the original MDPs model have lower performance under the new MDPs model.

\section{CONCLUSION}
In this paper, we propose DEA-RRL, a fully decoupled safe reinforcement learning framework.
The decoupling of task performance and safety is achieved by pre-training the recovery policies that maximize safety.
The task policies are free to explore without considering safety at all, and safety in exploration is achieved by combining recovery policies with behavior correction.
We show that, our approach is able to identify the dead-end states in determined MDPs, which is the maximum range that a policy can explore safely.
Using appropriate safety thresholds to prevent policies from going into dead-ends can ensure safety without compromising reward improvement.
We demonstrate in a series of experiments that our approach balances task performance and safety well, suggesting that our approach has potential to be used in realistic environmental tasks.

Future work will consider the following aspects:
(1) introducing model-based reinforcement learning to estimate uncertainty, allowing $\epsilon_{safe}$ to be adaptive,  and
(2) obtaining an offline safe reinforcement learning algorithm that involves behavior correction mechanism in offline training process of $\pi_{task}$ to improve task performance.

\bibliographystyle{IEEEtran}
\bibliography{IEEEabrv, paper}

\begin{thebibliography}{10}
\providecommand{\url}[1]{#1}
\csname url@rmstyle\endcsname
\providecommand{\newblock}{\relax}
\providecommand{\bibinfo}[2]{#2}
\providecommand\BIBentrySTDinterwordspacing{\spaceskip=0pt\relax}
\providecommand\BIBentryALTinterwordstretchfactor{4}
\providecommand\BIBentryALTinterwordspacing{\spaceskip=\fontdimen2\font plus
\BIBentryALTinterwordstretchfactor\fontdimen3\font minus
  \fontdimen4\font\relax}
\providecommand\BIBforeignlanguage[2]{{%
\expandafter\ifx\csname l@#1\endcsname\relax
\typeout{** WARNING: IEEEtran.bst: No hyphenation pattern has been}%
\typeout{** loaded for the language `#1'. Using the pattern for}%
\typeout{** the default language instead.}%
\else
\language=\csname l@#1\endcsname
\fi
#2}}

\bibitem{mnih2015human}
V.~Mnih, K.~Kavukcuoglu, D.~Silver, A.~A. Rusu, J.~Veness, M.~G. Bellemare,
  A.~Graves, M.~Riedmiller, A.~K. Fidjeland, G.~Ostrovski, \emph{et~al.},
  ``Human-level control through deep reinforcement learning,'' \emph{nature},
  vol. 518, no. 7540, pp. 529--533, 2015.

\bibitem{kendall2019learning}
A.~Kendall, J.~Hawke, D.~Janz, P.~Mazur, D.~Reda, J.-M. Allen, V.-D. Lam,
  A.~Bewley, and A.~Shah, ``Learning to drive in a day,'' in \emph{2019
  International Conference on Robotics and Automation (ICRA)}.\hskip 1em plus
  0.5em minus 0.4em\relax IEEE, 2019, pp. 8248--8254.

\bibitem{bodnar2019quantile}
C.~Bodnar, A.~Li, K.~Hausman, P.~Pastor, and M.~Kalakrishnan, ``Quantile qt-opt
  for risk-aware vision-based robotic grasping,'' \emph{arXiv preprint
  arXiv:1910.02787}, 2019.

\bibitem{zhao2023state}
W.~Zhao, T.~He, R.~Chen, T.~Wei, and C.~Liu, ``State-wise safe reinforcement
  learning: A survey,'' \emph{arXiv preprint arXiv:2302.03122}, 2023.

\bibitem{thananjeyan2021recovery}
B.~Thananjeyan, A.~Balakrishna, S.~Nair, M.~Luo, K.~Srinivasan, M.~Hwang, J.~E.
  Gonzalez, J.~Ibarz, C.~Finn, and K.~Goldberg, ``Recovery rl: Safe
  reinforcement learning with learned recovery zones,'' \emph{IEEE Robotics and
  Automation Letters}, vol.~6, no.~3, pp. 4915--4922, 2021.

\bibitem{schafer2021decoupling}
L.~Sch{\"a}fer, F.~Christianos, J.~Hanna, and S.~V. Albrecht, ``Decoupling
  exploration and exploitation in reinforcement learning,'' in \emph{ICML 2021
  Workshop on Unsupervised Reinforcement Learning}, 2021.

\bibitem{ha2020learning}
S.~Ha, P.~Xu, Z.~Tan, S.~Levine, and J.~Tan, ``Learning to walk in the real
  world with minimal human effort,'' \emph{arXiv preprint arXiv:2002.08550},
  2020.

\bibitem{achiam2017constrained}
J.~Achiam, D.~Held, A.~Tamar, and P.~Abbeel, ``Constrained policy
  optimization,'' in \emph{International conference on machine learning}.\hskip
  1em plus 0.5em minus 0.4em\relax PMLR, 2017, pp. 22--31.

\bibitem{yang2020projection}
T.-Y. Yang, J.~Rosca, K.~Narasimhan, and P.~J. Ramadge, ``Projection-based
  constrained policy optimization,'' \emph{arXiv preprint arXiv:2010.03152},
  2020.

\bibitem{kim2022safety}
D.~Kim, Y.~Kim, K.~Lee, and S.~Oh, ``Safety guided policy optimization,'' in
  \emph{2022 IEEE/RSJ International Conference on Intelligent Robots and
  Systems (IROS)}.\hskip 1em plus 0.5em minus 0.4em\relax IEEE, 2022, pp.
  2462--2467.

\bibitem{cheng2019end}
R.~Cheng, G.~Orosz, R.~M. Murray, and J.~W. Burdick, ``End-to-end safe
  reinforcement learning through barrier functions for safety-critical
  continuous control tasks,'' in \emph{Proceedings of the AAAI conference on
  artificial intelligence}, vol.~33, no.~01, 2019, pp. 3387--3395.

\bibitem{shao2021reachability}
Y.~S. Shao, C.~Chen, S.~Kousik, and R.~Vasudevan, ``Reachability-based
  trajectory safeguard (rts): A safe and fast reinforcement learning safety
  layer for continuous control,'' \emph{IEEE Robotics and Automation Letters},
  vol.~6, no.~2, pp. 3663--3670, 2021.

\bibitem{wachi2020safe}
A.~Wachi and Y.~Sui, ``Safe reinforcement learning in constrained markov
  decision processes,'' in \emph{International Conference on Machine
  Learning}.\hskip 1em plus 0.5em minus 0.4em\relax PMLR, 2020, pp. 9797--9806.

\bibitem{luo2021mesa}
M.~Luo, A.~Balakrishna, B.~Thananjeyan, S.~Nair, J.~Ibarz, J.~Tan, C.~Finn,
  I.~Stoica, and K.~Goldberg, ``Mesa: Offline meta-rl for safe adaptation and
  fault tolerance,'' \emph{arXiv preprint arXiv:2112.03575}, 2021.

\bibitem{finn2017model}
C.~Finn, P.~Abbeel, and S.~Levine, ``Model-agnostic meta-learning for fast
  adaptation of deep networks,'' in \emph{International conference on machine
  learning}.\hskip 1em plus 0.5em minus 0.4em\relax PMLR, 2017, pp. 1126--1135.

\bibitem{whitney2021decoupled}
W.~F. Whitney, M.~Bloesch, J.~T. Springenberg, A.~Abdolmaleki, K.~Cho, and
  M.~Riedmiller, ``Decoupled exploration and exploitation policies for
  sample-efficient reinforcement learning,'' \emph{arXiv preprint
  arXiv:2101.09458}, 2021.

\bibitem{srinivasan2020learning}
K.~Srinivasan, B.~Eysenbach, S.~Ha, J.~Tan, and C.~Finn, ``Learning to be safe:
  Deep rl with a safety critic,'' \emph{arXiv preprint arXiv:2010.14603}, 2020.

\bibitem{zhang2022safety}
L.~Zhang, Z.~Yan, L.~Shen, S.~Li, X.~Wang, and D.~Tao, ``Safety correction from
  baseline: Towards the risk-aware policy in robotics via dual-agent
  reinforcement learning,'' in \emph{2022 IEEE/RSJ International Conference on
  Intelligent Robots and Systems (IROS)}.\hskip 1em plus 0.5em minus
  0.4em\relax IEEE, 2022, pp. 9027--9033.

\bibitem{fatemi2019dead}
M.~Fatemi, S.~Sharma, H.~Van~Seijen, and S.~E. Kahou, ``Dead-ends and secure
  exploration in reinforcement learning,'' in \emph{International Conference on
  Machine Learning}.\hskip 1em plus 0.5em minus 0.4em\relax PMLR, 2019, pp.
  1873--1881.

\bibitem{fatemi2021medical}
M.~Fatemi, T.~W. Killian, J.~Subramanian, and M.~Ghassemi, ``Medical dead-ends
  and learning to identify high-risk states and treatments,'' \emph{Advances in
  Neural Information Processing Systems}, vol.~34, pp. 4856--4870, 2021.

\bibitem{killian2023risk}
T.~W. Killian, S.~Parbhoo, and M.~Ghassemi, ``Risk sensitive dead-end
  identification in safety-critical offline reinforcement learning,''
  \emph{arXiv preprint arXiv:2301.05664}, 2023.

\bibitem{thomas2021safe}
G.~Thomas, Y.~Luo, and T.~Ma, ``Safe reinforcement learning by imagining the
  near future,'' \emph{Advances in Neural Information Processing Systems},
  vol.~34, pp. 13\,859--13\,869, 2021.

\bibitem{janner2019trust}
M.~Janner, J.~Fu, M.~Zhang, and S.~Levine, ``When to trust your model:
  Model-based policy optimization,'' \emph{Advances in neural information
  processing systems}, vol.~32, 2019.

\bibitem{altman1999constrained}
E.~Altman, \emph{Constrained Markov decision processes}.\hskip 1em plus 0.5em
  minus 0.4em\relax CRC press, 1999, vol.~7.

\bibitem{arulkumaran2017deep}
K.~Arulkumaran, M.~P. Deisenroth, M.~Brundage, and A.~A. Bharath, ``Deep
  reinforcement learning: A brief survey,'' \emph{IEEE Signal Processing
  Magazine}, vol.~34, no.~6, pp. 26--38, 2017.

\bibitem{kostrikov2021offline}
I.~Kostrikov, A.~Nair, and S.~Levine, ``Offline reinforcement learning with
  implicit q-learning,'' \emph{arXiv preprint arXiv:2110.06169}, 2021.

\bibitem{haarnoja2018soft}
T.~Haarnoja, A.~Zhou, P.~Abbeel, and S.~Levine, ``Soft actor-critic: Off-policy
  maximum entropy deep reinforcement learning with a stochastic actor,'' in
  \emph{International conference on machine learning}.\hskip 1em plus 0.5em
  minus 0.4em\relax PMLR, 2018, pp. 1861--1870.

\bibitem{ray2019benchmarking}
A.~Ray, J.~Achiam, and D.~Amodei, ``Benchmarking safe exploration in deep
  reinforcement learning,'' \emph{arXiv preprint arXiv:1910.01708}, vol.~7,
  no.~1, p.~2, 2019.

\bibitem{yang2021wcsac}
Q.~Yang, T.~D. Sim{\~a}o, S.~H. Tindemans, and M.~T. Spaan, ``Wcsac: Worst-case
  soft actor critic for safety-constrained reinforcement learning,'' in
  \emph{Proceedings of the AAAI Conference on Artificial Intelligence},
  vol.~35, no.~12, 2021, pp. 10\,639--10\,646.

\bibitem{feng2023dense}
S.~Feng, H.~Sun, X.~Yan, H.~Zhu, Z.~Zou, S.~Shen, and H.~X. Liu, ``Dense
  reinforcement learning for safety validation of autonomous vehicles,''
  \emph{Nature}, vol. 615, no. 7953, pp. 620--627, 2023.

\end{thebibliography}

\end{document}